# Approximate Principal Direction Trees

Mark McCartin-Lim                                              MARKML@CS.UMASS.EDU
Andrew McGregor                                              MCGREGOR@CS.UMASS.EDU
Rui Wang                                                      RUIWANG@CS.UMASS.EDU
University of Massachusetts Amherst, 140 Governors Drive, Amherst, MA 01002-9264 USA

## Abstract

We introduce a new spatial data structure for high dimensional data called the *approximate principal direction tree* (APD tree) that adapts to the intrinsic dimension of the data. Our algorithm ensures vector-quantization accuracy similar to that of computationally-expensive PCA trees with similar time-complexity to that of lower-accuracy RP trees.

APD trees use a small number of power-method iterations to find splitting planes for recursively partitioning the data. As such they provide a natural trade-off between the running-time and accuracy achieved by RP and PCA trees. Our theoretical results establish a) strong performance guarantees regardless of the convergence rate of the power-method and b) that $O(\log d)$ iterations suffice to establish the guarantee of PCA trees when the intrinsic dimension is $d$. We demonstrate this trade-off and the efficacy of our data structure on both the CPU and GPU.

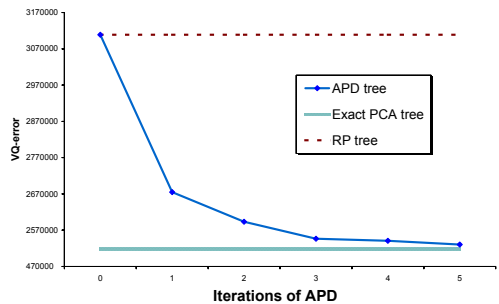

*Figure 1.* **Convergence of APD Trees to PCA Trees on the MNIST Dataset (depth 4 trees):** The VQ error achieved by using APD trees with only a few power-method iterations is close to that of PCA trees. This substantially improves upon the error for RP trees without significant computational overhead.

## 1. Introduction

*Spatial partition trees* are data structures that hierarchically subdivide a set of data points with the goal that resulting partitions contain "similar" points. A ubiquitous example is the $k$-d tree, which is widely used in many unsupervised learning methods including classification, regression, nearest-neighbor finding, and vector quantization.

At level $L$ of the tree, the data has been partitioned into $2^L$ different classes. Every time we subdivide a set of points, we do so with the hope of reducing the average distance between points in the same subsets as much as possible. The goal is similar to that in *k-means clustering*, but unfortunately, *k*-means clustering is an NP-hard problem even when $k = 2$ (Dasgupta, 2008).

We know that $k$-d trees are vulnerable to the so-called *curse of dimensionality* – if the dimensionality of our data is $D$, then we may have to traverse $O(D)$ levels to halve the average diameter. This is shown to be true even when the *intrinsic dimensionality* of the data is low (Verma et al., 2009). Thus, $k$-d tree are not well suited to applications that involve high-dimensional data, because to produce a good classifier, we would need to branch our tree $O(2^D)$ times. Similarly, it has been shown that using a $k$-d tree to do nearest-neighbor queries in high-dimensional data is not much more efficient than scanning every element (Lee & Wong, 1977).

Recently, random-projection (RP) trees (Dasgupta & Freund, 2008) and PCA trees (Verma et al., 2009) have been proposed as alternatives to $k$-d trees that adapt to intrinsic dimensionality. RP trees and PCA trees differ from $k$-d trees in that their splitting planes are not restricted to being axis-aligned. Thus, unlike $k$-d trees, they can adapt to the covariance of the data (see Figure 2).

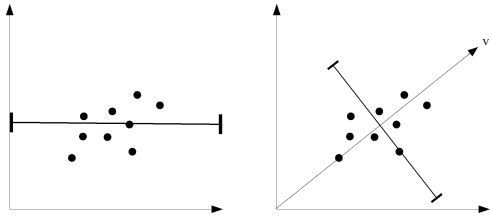

Figure 2. Left: $k$-d tree – partitions with hyperplane perpendicular to an axis. Right: PCA tree – partitions with hyperplane perpendicular to the principal direction $v$.

Of the two choices, RP trees are appealing because they can be computed very efficiently. Each subdivision in an RP tree is determined by a randomly-chosen hyperplane. PCA trees are significantly more expensive to compute, because the normal of each hyperplane is chosen by doing principal component analysis (PCA) on the data to find the principal direction. This requires one to compute the covariance matrix and perform eigendecomposition, or to do singular value decomposition (SVD). However, PCA trees perform significantly better than RP trees at reducing the average diameter, as demonstrated in (Verma et al., 2009).

**APD Trees.** We propose a new spatial partition tree, the *approximate principal direction tree* or APD trees, which generalize the idea behind RP trees, but perform almost as well as PCA trees when it comes to reducing average diameter with respect to intrinsic dimensionality.

When choosing the normal of the hyperplane, APD trees start with random vectors like RP trees, but then apply a small number of power-method iterations (Burden & Faires, 2010) to these vectors. The power method is often used in data intensive applications to approximate principal eigenvectors (e.g., Google's PageRank (Wills, 2007)) and for spectral clustering (Lin & Cohen, 2010). However, it is important to note that in our application we are not simply concerned with getting a good approximation of the principal eigenvector. Rather, we are finding a hyperplane that will yield a good subdivision. This is important because the power method can be slow to converge when the first principal component has the same variance as the second principal component. However, this is not a concern for us since either component (or some linear combination thereof) would suffice for a good split.

We show strong performance guarantees with only one or two power-method iterations. See Figure 1 for an empirical illustration. Furthermore, when the intrinsic dimension of a dataset is $d$, we prove that $O(\log d)$ power-iterations are sufficient to produce trees that reduce average diameter at the same rate as PCA trees. Again, this is true even when the principal direction is not dominant in the data.

**Outline.** In Section 2, we present the necessary definitions including that of *local covariance dimension* (Dasgupta & Freund, 2008) and what it means for a tree to adapt to intrinsic dimensionality. In Section 3, we presents the algorithm for building APD trees. Then, in Section 4, we prove that it adapts to intrinsic dimensionality with the average diameter converging at a similar rate to PCA trees. This is further demonstrated by our experimental results (Section 5), which also show that APD tree is much faster than the standard PCA tree algorithm. Finally, we also present a GPU-based implementation, which is even faster.

## 2. Preliminary Definitions

### 2.1. Average Diameter

For a given set of points $S = \{x_1, \ldots, x_n\} \subset \mathbb{R}^D$ we measure their similarity in terms of the average distance between points in the set. We will use the following notation:

**Definition 2.1** (Average diameter of a set of points).

$$\Delta_a(S) = \frac{1}{|S|}\sqrt{\sum_{x,y \in S} \|x - y\|^2}$$

or equivalently[1]

$$\Delta_a(S) = \sqrt{\frac{2}{|S|} \sum_{x \in S} \|x - \text{mean}(S)\|^2} \ ,$$

where $\text{mean}(S) = \sum_{x \in S} x/|S|$.

We are interested in partitioning $S$ as $\{S_1, S_2\}$ such that the average diameter of $S_1$ and $S_2$ is small.

**Definition 2.2** (Average diameter over two sets).

$$\Delta_a(S_1, S_2) = \sqrt{\frac{\Delta_a^2(S_1)|S_1| + \Delta_a^2(S_2)|S_2|}{|S_1| + |S_2|}} \ .$$

### 2.2. Local Covariance Dimension

Several possible definitions of *intrinsic dimension* are discussed in (Verma et al., 2009). The one they use in their analysis of RP trees and PCA trees, *local covariance dimension* is based upon a statistical representation of the data, and is thus well-suited to modeling data from machine learning problems. We will analysis APD trees using the same definition.

---
[1] See, e.g., (Dasgupta & Freund, 2008).

Recall that the covariance matrix $C \in \mathbb{R}^{D \times D}$ of a set of points $S = \{x_1, \ldots, x_n\} \subset \mathbb{R}^D$ with $\text{mean}(S) = 0$, can be written as $C = X^T X$ where $X \in \mathbb{R}^{n \times D}$ is the matrix whose $i$-th row equals $x_i$. The idea behind local covariance dimension is as follows. The eigenvectors of the covariance matrix form an orthonormal basis for the data. If a small number of the eigenvectors describe almost all the variance in the data, then we know that the data is well represented by the subspace spanned by those eigenvectors. We can think of the dimension of this subspace as being the intrinsic dimension of the data.

**Definition 2.3** (Local Covariance Dimension). *Let $\lambda_1 \geq \lambda_2 \geq \ldots \geq \lambda_D$ be the eigenvalues of the covariance matrix of $S = \{x_1, \ldots, x_n\} \subset \mathbb{R}^D$. We say $S$ has local covariance dimension $(d, \varepsilon)$ for $d \leq D$ and $0 < \varepsilon < 1$ when*

$$\sum_{i=1}^d \lambda_i \geq (1-\varepsilon) \sum_{i=1}^D \lambda_i \ .$$

Notice that the local covariance dimension is also parametrized by $\varepsilon$, which corresponds to how closely the subspace represents the data.

We say a method for partitioning $S$ *adapts to the intrinsic dimension* of $S$ if the resulting partition $\{S_1, S_2\}$ satisfies $\Delta_a(S_1, S_2) < f(S) \cdot \Delta_a(S)$ where $f$ is independent of $D$, the extrinsic dimension.

## 3. The APD Construction

A spatial partition tree is determined by the rule used to partition the points at each node of the tree. Before we discuss the partitioning rule used in APD trees, we review a general template for possible rules. This template is presented in Algorithm 1. If we assume $S$ corresponds to the set of points assigned to a particular node in the tree, then it is obvious that this meta-algorithm can be applied recursively to produce a balanced binary tree.

The tree produced by this meta-algorithm is a hybrid of a *BSP tree* and a *sphere tree* (Devroye et al., 1996). If $S$ is considered to contain "outliers", we use a sphere to partition points that are close to the center of the data away from those that are not. We discuss this case further in Section 3.2. Otherwise, we use a hyperplane to partition the data. The choice of this hyper-plane depends upon a *splitting rule*, which determines the normal for the hyperplane. Two existing splitting rules are the RP and PCA rules:

1. **RP rule:** $p$ is chosen uniformly at random from the unit sphere in $\mathbb{R}^D$.

2. **PCA rule:** $p$ is the principal eigenvector of $C$.

**Algorithm 1** Tree construction meta-algorithm
  **if** $S$ has outliers **then**
    $D := \{\|x - \text{mean}(S)\| \mid x \in S\}$
    $S_1 := \{x \in S \mid \|x - \text{mean}(S)\| \leq \text{median}(D)\}$
  **else**
    Choose $p \in \mathbb{R}^D$ according to a *splitting rule*
    $P := \{x \cdot p \mid x \in S\}$
    $S_1 := \{x \in S \mid x \cdot p \leq \text{median}(P)\}$
  **end if**
  $S_2 := S \setminus S_1$
  **return** $\{S_1, S_2\}$

**Algorithm 2** APD splitting rule
  $p :=$ a random vector $s \in \mathbb{R}^D$
  **for** $1 \ldots t$ **do**
    $q := \sum_{h=1}^n (x_h \cdot p) \, x_h^T$
    $p := q/\|q\|$
  **end for**
  **return** $p$

As we mentioned earlier, the downside of the PCA splitting rule is that computing $p$ either requires one to compute the covariance matrix and perform eigendecomposition, or to do singular value decomposition on the data. Both are computationally intensive tasks. On the other hand, applying the RP splitting rule is computationally trivial. However, this splitting rule does not achieve as good accuracy as the PCA splitting rule.

### 3.1. The New Splitting Rule

The new splitting rule we propose allows us to achieve similar accuracy to that achieved by the PCA rule without the computational overhead. The rule is based on the Power Method (Burden & Faires, 2010), a well-known technique for approximating eigenvectors. See Algorithm 2. The technique translates nicely into a parallel algorithm we can implement on the GPU, as we will see in Section 5.

**Theorem 3.1.** *For $i \in [D]$, let $p_i \in \mathbb{R}^D$ and $\lambda_i > 0$ be the (normalized) eigenvectors and eigenvalues of the covariance matrix $C$. If $s = \sum_{i=1}^D \beta_i p_i$ is the initial* vector used in the APD splitting rule, then the vector returned by the splitting rule is

$$p = \frac{C^t s}{\|C^t s\|} = \frac{\sum_{i=1}^D \lambda_i^t \beta_i p_i}{\sqrt{\sum_{j=1}^D \lambda_j^{2t} \beta_j^2}} .$$

*Proof.* The result follows from the observation that $\sum_{h=1}^n (x_h \cdot p) x_h^T = (X^T X) p = Cp$. □

Below are two properties of the APD splitting rule:

**Generalization of RP and PCA:** As $t$ increases, $p$ will converge to the principal eigenvector of the covariance matrix, i.e., the vector we would get if we were to use the PD splitting rule. On the other hand, when $t = 0$, $p$ is a random unit vector in $\mathbb{R}^D$, and thus equivalent to if we were using the RP splitting rule. So intuitively, using the APD splitting rule gives us a trade-off between the RP splitting rule and the PCA splitting rule.

**Doesn't require convergence of power method:** It is important to emphasize that the idea behind applying the power method in our setting is not to approximate the principal eigenvector per se. We need the splitting rule to be computed quickly and will primarily be interested in the case when $t = 1$ or 2. In this case, it is unlikely that the vector returned is similar to the principal eigenvector. For example, the power method converges slowly when the first few principal values are very close to each other. However, this is not an issue in our setting since a few iterations will still ensure that the direction of $p$ has high variance.

### 3.2. Fast outlier detection

The tree construction meta-algorithm has a special case for they $S$ contains *outliers*, i.e., when there average distance between points is significantly less than the maximum distance between two points. This is important, because even if we were to find a good hyperplane to split our data, the average diameter of the resulting partitions will still be influenced by the outliers.

Following (Dasgupta & Freund, 2008; Verma et al., 2009), we say that $S$ has outliers if the maximum depends on the relative size of the average diameter and the maximum diameter:

**Definition 3.1** (Outliers). *For an user-definable parameter $c > 0$, we say $S$ contains outliers if*

$$\Delta^2(S) > c \Delta_a^2(S)$$

*where $\Delta(S) = \max_{x,y \in S} \|x - y\|$.*

It was shown in (Dasgupta & Freund, 2008) that in the case $S$ has outliers according to this definition, the meta-algorithm (Algorithm 1) still guarantees a constant reduction in average diameter.

**Proposition 3.1.** *Suppose $\Delta^2(S) > c\Delta_a^2(S)$, so that $S$ is split into $\{S_1, S_2\}$ as described in Algorithm 1. Then the following holds:*

$$\Delta_a^2(S_1, S_2) \leq \left(\frac{1}{2} + \frac{2}{c}\right) \Delta_a^2(S)$$

Unfortunately, it is computationally expensive to use Definition 3.1 to determine if there are outliers, because calculating $\Delta^2(S)$ involves comparing $O(|S|^2)$ distances. Instead we proposed a simple variant. For an arbitrary point $a \in S$, let $D(S) = \max_{x \in S} \|x - a\|$. It can easily be shown that $\Delta(S)/2 \leq D(S) \leq \Delta(S)$.

**Definition 3.2** (Outlier heuristic). *$S$ has outliers if*

$$D^2(S) > c\Delta_a^2(S) .$$

### 3.3. Comparison between splitting rules

We can now state our main theoretical result for the APD splitting rule and contrast it with the analogous results for RP and and PCA trees. Henceforth, we assume that there are no outliers, i.e., $\Delta^2(S) < c\Delta_a^2(S)$.

The following results were shown in (Dasgupta & Freund, 2008) and (Verma et al., 2009).

**Proposition 3.2.** *There exist constants $c_1, c_2 \in (0, 1)$ such that if $S$ has local covariance dimension $(d, c_1)$:*

1. **RP rule:** *If $p$ is chosen uniformly at random from the unit sphere in $\mathbb{R}^D$ then,*

$$\mathbb{E}\left[\Delta_a^2(S_1, S_2)\right] < (1 - c_2/d) \Delta_a^2(S) .$$

2. **PCA rule:** *$p$ is the principal eigenvector then,*

$$\Delta_a^2(S_1, S_2) < \left(1 - c_2/k^2\right) \Delta_a^2(S) ,$$

*where $k = \frac{1}{\lambda_1} \sum_{i=1}^d \lambda_i$. Note that $k \leq d$.[2]*

These results show that RP trees and PCA trees both adapt to intrinsic dimensionality, since the squared average diameter of nodes in the tree is decreasing as a function of $d$, the local covariance dimension of the data, rather than its extrinsic dimension $D$.

In the next section we will prove that a similar diameter reduction guarantee holds for APD trees:

---
[2] $k$ can be much less than $d$. For example, in the MNIST data set $k^2 < d$ for $d$ larger than 80.

**Theorem 3.2** (Main Result). *For $c_1, \delta \in (0,1)$, there exists constant[3] $c_2 \in (0,1)$ such that if*

1. *$S$ has local covariance dimension $(d, c_1)$ and*

2. *$p \in \mathbb{R}^D$ is returned by the APD splitting rule with $t$ iterations*

*then with probability $1 - \delta$,*

$$\Delta_a^2(S_1, S_2) < \left(1 - \frac{c_2}{k^2(d-1)^{\frac{2}{2t+1}}}\right) \Delta_a^2(S)$$

*where $k = \frac{1}{\lambda_1} \sum_{i=1}^{d} \lambda_i$.*

Note that for $t = O(\log d)$, this gives

$$\Delta_a^2(S_1, S_2) < \left(1 - c_2/k^2\right) \Delta_a^2(S) ,$$

i.e., the same improvement as that achieved for PCA trees. However, even for smaller $t$, the bound only has a weak dependence on $d$.

## 4. Theoretical Analysis of APD trees

To prove Theorem 3.2 we will need to analyze the quantity

$$V(S, p) = \frac{1}{n} \sum_{x \in S} (x \cdot p)^2 .$$

For a fixed vector $p$, observe that $V(S, p)$ corresponds to the variance of $x \cdot p$ when $x$ is drawn uniformly at random from $S$. Intuitively, it makes sense that a good splitting vector is one for which $V(S, p)$ is large. Specifically, if we can prove a lower bound for $V(S, p)$ when $p$ is chosen according to the APD splitting rule, then we can appeal to the following variant of a proposition from (Verma et al., 2009).

**Proposition 4.1.** *There exist constants $0 < c_1, c_2 < 1$ with the following property. Suppose $\Delta^2(S) \leq c\Delta_a^2(S)$, so that $S$ is split into $\{S_1, S_2\}$ using the projection vector $p$. If $S$ has local covariance dimension $(d, c_1)$, then:*

$$\Delta_a^2(S_1, S_2) < \left(1 - \frac{c_2}{k^2} \left(\frac{V(S,p)}{\lambda_1}\right)^2\right) \Delta_a^2(S)$$

*where $k = \frac{1}{\lambda_1} \sum_{i=1}^{d} \lambda_i$.*

To lower bound $V(S, p)$ we prove the following sequence of lemmas that relate $V(S, p)$ to the eigenvectors $p_1, \ldots, p_D$ and corresponding eigenvalues $\lambda_1, \ldots, \lambda_D$ of the covariance matrix $C$. Recall that $\lambda_1 \geq \lambda_2 \geq \ldots \geq \lambda_D$ and that the eigenvalues are non-negative since $C$ is positive semi-definite.

---
[3]Where $c_2 \propto c_\delta^2 (1-c_1)^4$ and $c_\delta$ is at worst polynomial in $\delta$ and the corresponding percentile of the $\chi^2$ distribution. Empirically, it suffices for $c_\delta = 1/3$ when $\delta = 0.01$.

**Lemma 4.1.** *For any $q = \sum_{i=1}^{D} \alpha_i p_i$,*

$$V(S, q) = \sum_{i=1}^{D} \lambda_i \alpha_i^2 .$$

*Proof.*

$$V(S, q) = \sum_{i=1}^{n} \frac{(x_i \cdot q)^2}{n} = \frac{1}{n} (Xq)^T (Xq) = q^T C q$$

$$= \left(\sum_{i=1}^{D} \alpha_i p_i\right)^T \left(\sum_{i=1}^{D} C \alpha_i p_i\right)$$

$$= \left(\sum_{i=1}^{D} \alpha_i p_i\right)^T \left(\sum_{i=1}^{D} \lambda_i \alpha_i p_i\right) = \sum_{i=1}^{D} \lambda_i \alpha_i^2 ,$$

where the last equality follows because the eigenvectors are orthonormal. □

Lemma 4.1 and Theorem 3.1 imply the next lemma.

**Lemma 4.2.** *Let $p$ be the vector computed by the APD rule after $t$ iterations where $s = \sum_i \beta_i p_i$ is the initial vector. Then*

$$V(S, p) = \frac{\sum_{i=1}^{D} \lambda_i^{2t+1} \beta_i^2}{\sum_{j=1}^{D} \lambda_j^{2t} \beta_j^2} .$$

We now analyze the distribution of the $\beta_i$ coefficients to prove the next lemma.

**Lemma 4.3.** *For any $\delta < 1$ and $\lambda_1, \ldots, \lambda_D \geq 0$, there exists $c_\delta > 0$ with*

$$\mathbb{P}\left[\frac{\sum_{i=1}^{D} \lambda_i^{2t+1} \beta_i^2}{\sum_{j=1}^{D} \lambda_j^{2t} \beta_j^2} \geq c_\delta \frac{\sum_{i=1}^{D} \lambda_i^{2t+1}}{\sum_{j=1}^{D} \lambda_j^{2t}}\right] \geq 1 - \delta , \quad (1)$$

*if the direction of $s = \sum_i \beta_i p_i$ is chosen uniformly.*

*Proof.* Let $\{\gamma_i\}_{i \in [D]}$ be independently distributed $\chi^2$-random variables with one degree of freedom. Then, since the direction of the vector $\langle \beta_1, \ldots, \beta_D \rangle$ is chosen uniformly at random in the APD rule, we know that

$$\frac{\sum_{i=1}^{D} \lambda_i^{2t+1} \beta_i^2}{\sum_{i=1}^{D} \lambda_i^{2t} \beta_i^2} \sim \frac{\sum_{i=1}^{D} \lambda_i^{2t+1} \gamma_i}{\sum_{i=1}^{D} \lambda_i^{2t} \gamma_i} = \frac{\sum_{i=1}^{D} \mu_i^{2t+1} \gamma_i}{\sum_{i=1}^{D} \mu_i^{2t} \gamma_i} ,$$

where $\mu_i = \lambda_i/\lambda_1$ so that $1 = \mu_1 \geq \mu_2 \geq \cdots \geq \mu_D$. This follows because a) a random point on the unit sphere can be sampled by choosing each coefficient according to the standard normal distribution and then renormalizing (Muller, 1958); and b) the square of a variable with standard normal distribution has the $\chi^2$-distribution with one degree of freedom.

The lemma follows by the union bound if we can show that there exist constants $c_1, c_2 > 0$ such that:

$$\mathbb{P}\left[\sum_{j=1}^{D} \mu_j^{2t} \gamma_j \geq c_1 \sum_{j=1}^{D} \mu_j^{2t}\right] \leq \delta/2 \quad (2)$$

$$\mathbb{P}\left[\sum_{i=1}^{D} \mu_i^{2t+1} \gamma_i \leq c_2 \sum_{i=1}^{D} \mu_i^{2t+1}\right] \leq \delta/2 \quad (3)$$

since then Eq. 1 holds with $c_\delta = c_2/c_1$.

For the first inequality, note that $\mathbb{E}[\gamma_i] = 1$ (expectation of $\chi^2$ distribution) and hence

$$\mathbb{E}\left[\sum_{j=1}^{D} \mu_j^{2t} \gamma_j\right] = \sum_{j=1}^{D} \mu_j^{2t} .$$

Therefore Eq. 2 follows from an application of the Markov inequality with $c_1 = 2/\delta$.

For the second inequality there are two cases. First suppose that $\sum_{i=1}^{D} \mu_i^{2t+1} \leq 16/\delta$. Using the inverse CDF of the $\chi^2$ distribution, we can compute $\sigma$ where $\mathbb{P}[\gamma_1 \leq \sigma] = \delta/2$. Then note that,

$$\sum_{i=1}^{D} \mu_i^{2t+1} \gamma_i \geq \gamma_1 \geq \sigma \geq \frac{\delta\sigma}{16} \sum_{i=1}^{D} \mu_i^{2t+1}$$

where the second inequality ($\gamma_1 \geq \sigma$) holds with probability at least $1 - \delta/2$.

Alternatively, suppose that $\sum_{i=1}^{D} \mu_i^{2t+1} \geq 16/\delta$. Then, appealing to the Chebyshev inequality given that

$$\mathbb{V}\left[\sum_{i=1}^{D} \mu_i^{2t+1} \gamma_i\right] = 2 \sum_{i=1}^{D} \mu_i^{4t+2} \leq 2 \sum_{i=1}^{D} \mu_i^{2t+1},$$

we conclude

$$\mathbb{P}\left[\sum_{i=1}^{D} \mu_i^{2t+1} \gamma_i \leq \frac{1}{2} \sum_{i=1}^{D} \mu_i^{2t+1}\right] \leq \frac{8 \sum_{i=1}^{D} \mu_i^{2t+1}}{(\sum_{i=1}^{D} \mu_i^{2t+1})^2} \leq \frac{\delta}{2} .$$

This establishes Eq. 3 with $c_2 = \min(\delta\sigma/16, 1/2)$. □

**Lemma 4.4.** *For $S = \{x_1, \ldots, x_n\} \in \mathbb{R}^D$ with local covariance dimension $(d, \varepsilon)$ with $\text{mean}(S) = 0$.*

$$\frac{\sum_{i=1}^{D} \lambda_i^{2t+1}}{\sum_{j=1}^{D} \lambda_j^{2t}} \geq (1 - \varepsilon) \frac{\sum_{i=1}^{d} \lambda_i^{2t+1}}{\sum_{j=1}^{d} \lambda_j^{2t}} .$$

*Proof.* Since $\sum_{i=1}^{D} \lambda_i^{2t+1} \geq \sum_{i=1}^{d} \lambda_i^{2t+1}$, it suffices to show that $\sum_{j=1}^{d} \lambda_j^{2t} \geq (1 - \varepsilon) \sum_{j=1}^{D} \lambda_j^{2t}$.

From Definition 2.3, it follows that $\sum_{j=1}^{d} \lambda_j \geq (1-\varepsilon) \sum_{j=1}^{D} \lambda_j$ and therefore $\varepsilon \sum_{j=1}^{d} \lambda_j \geq (1-\varepsilon) \sum_{j=d+1}^{D} \lambda_j$. And hence, it follows that:

$$\varepsilon \sum_{j=1}^{d} \lambda_j^{2t} \geq \varepsilon \sum_{j=1}^{d} \lambda_j \lambda_d^{2t-1} \geq (1-\varepsilon) \sum_{j=d+1}^{D} \lambda_j \lambda_d^{2t-1}$$

$$\geq (1-\varepsilon) \sum_{j=d+1}^{D} \lambda_j^{2t}$$

and therefore $\sum_{j=1}^{d} \lambda_j^{2t} \geq (1-\varepsilon) \sum_{j=1}^{D} \lambda_j^{2t}$. □

**Theorem 4.1.** *Let $p$ be the vector computed by the APD rule after $t \geq 1$ iterations. Then with probability $1 - \delta$, there exists a constant $c_\delta > 0$ such that*

$$V(S, p) \geq \lambda_1 c_\delta (1-\varepsilon) \frac{1 + 2t}{1 + 2t + 2t^{\frac{2t}{2t+1}} (d-1)^{\frac{1}{2t+1}}} .$$

*Note this implies $V(S, p) \geq \lambda_1 c_\delta (1-\varepsilon)(d-1)^{\frac{-1}{2t+1}}/2$.*

*Proof.* Let $\alpha = 1/(2t)$. By Lemmas 4.2, 4.3, and 4.4, if we set $\mu_i = \lambda_i^{2t}/\lambda_1^{2t}$, we get

$$\frac{V(S, p)}{c_\delta(1-\varepsilon)} \geq \frac{\sum_{i=1}^{d} \lambda_i^{2t+1}}{\sum_{j=1}^{d} \lambda_i^{2t}}$$

$$= \frac{\sum_{i=1}^{d} \mu_i^{1+\alpha}}{\sum_{j=1}^{d} \mu_i}$$

$$\geq \min_{1 \geq \mu_2, \ldots, \mu_d} \frac{1 + \sum_{i=2}^{d} \mu_i^{1+\alpha}}{1 + \sum_{j=2}^{d} \mu_i}$$

$$\geq \min_{1 \geq \mu_2, \ldots, \mu_d \geq 0} \frac{1 + (d-1) \sum_{j=2}^{d} (\frac{\mu_i}{d-1})^{1+\alpha}}{1 + \sum_{j=2}^{d} \mu_i}$$

$$\geq \min_{\mu \geq 0} \frac{1 + (d-1)^{-\alpha} \mu^{1+\alpha}}{1 + \mu}$$

$$\geq \min_{\mu \geq 0} \frac{1 + (d-1)^{-\alpha} \mu^{1+\alpha}}{1 + (d-1)^{-\alpha} \mu^{1+\alpha} + \mu} ,$$

where the third last inequality follows from the convexity of the function $x^{1+1/(2t)}$ and the second last inequality follows by setting $\mu = \sum_{j=2}^{d} \mu_i$.

By analyzing the derivatives, it can be verified that

$$f(u) = \frac{1 + \mu^{1+\alpha}/(d-1)^\alpha}{1 + \mu^{1+\alpha}/(d-1)^\alpha + \mu}$$

has a unique minimum at $u = \left((2t)^{2t}(d-1)\right)^{1/(2t+1)}$. Substituting in this value establishes the theorem. □

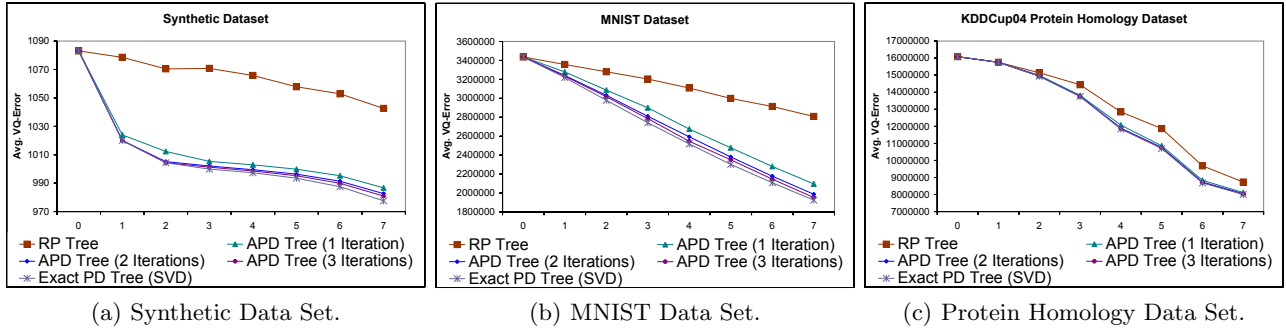

(a) Synthetic Data Set.  (b) MNIST Data Set.  (c) Protein Homology Data Set.

*Figure 3.* Compare VQ-error of RP, PCA and APD trees on the synthetic, MNIST, and Protein homology data sets. Each graph plots the decay of the VQ-errors (averaged over 15 runs) as the depth of the tree increases. For APD trees, we show results for one, two, and three power iterations. RP tree is equivalent to using no power iteration. Note that the differences between APD trees and PD trees are small, and using 1 iteration already provides very good quality.

*Table 1.* Timing of our CPU and GPU APD tree implementations on three data sets. $t$ denotes the number of power iterations ($t = 0$ is equivalent to RP). In all cases, the tree is subdivided to depth 4. The GPU implementation generally achieve $5 \sim 12\times$ speedup over the CPU version, both of which are written in MATLAB. Note that each additional power iteration incurs only small extra cost. We also show the timing of a CPU PCA tree implementation for comparison.

| Data set | | CPU timing (seconds) | | | | | | GPU timing (seconds) | | | | |
|---|---|---|---|---|---|---|---|---|---|---|---|---|
| Name | Size | t=0 | t=1 | t=2 | t=3 | t=4 | PCA | t=0 | t=1 | t=2 | t=3 | t=4 |
| Synthetic | $10K \times 1K$ | 4.88 | 4.94 | 5.00 | 5.06 | 5.12 | 22.6 | 0.41 | 0.43 | 0.45 | 0.47 | 0.49 |
| MINIST10K | $10K \times 784$ | 3.58 | 3.62 | 3.66 | 3.70 | 3.75 | 12.9 | 0.31 | 0.33 | 0.35 | 0.36 | 0.39 |
| KDDCup04 | $285K \times 74$ | 5.16 | 5.18 | 5.29 | 5.44 | 5.60 | 33.0 | 1.01 | 1.26 | 1.49 | 1.72 | 1.95 |

## 5. Experimental Results

We compare the quality of APD trees to that of RP trees and PCA trees by measuring the *vector quantization error* (VQ-error). In vector quantization, the goal is to map all vectors (or points) in a given data set to a small number of representative vectors (or points). This can be done with a spatial partitioning tree – the points belonging to each partition are represented by the average of the points in that partition. Following this, the VQ-error is defined as the average squared representation error. Specifically, if $S_1, S_2, \ldots, S_{2^\ell}$ are the sets of points associated with the leaves of a tree $T$ of depth $\ell$, then

$$VQ_T(S) = \sum_{i=1}^{2^\ell} \sum_{x \in S_i} \frac{\|x - \text{mean}(S_i)\|^2}{|S|} = \sum_{i=1}^{2^\ell} \frac{|S_i| \Delta_a^2(S_i)}{2|S|}.$$

We try to closely replicate the experiments done in (Freund et al., 2008), using the same kind of datasets and the same parameters. We ran our experiments on a synthetic dataset and the MNIST test dataset. Additionally we use a protein homology dataset from the KDD Cup 2004 data mining competition.

As in (Freund et al., 2008), the synthetic dataset consists of 10,000 points, each a 1,000-d vector and generated as follows: choose a peak value $p$ uniformly randomly from $[0, 1]$, and then generates the coordinates of the point from the normal distribution $N(p, 1)$. The MNIST test dataset is a set of 10,000 images of handwritten digits, each of which has been normalized to $28 \times 28$ pixels, and is thus a 784-d vector. The protein homology dataset consists of pairs of proteins that have been tested for homology. It contains 285,409 data points, each of which is a 74-d vector.

For RP trees and APD trees, we perform 15 runs of each experiment, and calculated the average VQ-error. For PCA trees, we only needed to run each experiment once since there is no randomization in that algorithm. The results are shown in Figure 3. Each curve plots the decay of average VQ-errors as the depth of the tree increases. For APD trees, we show results of applying 1, 2, and 3 power iterations. RP tree is equivalent to applying no power iteration. The quality differences between APD trees and PCA trees are small, while the differences of RP trees with them are much more visible. Note that using 1 power iteration in APD trees already provides very good quality and gives the the biggest drop in VQ-error from RP trees.

**GPU Implementation of APD Trees.** Our APD tree algorithm is well-suited for parallel computation on modern GPUs. Since it only relies on basic matrix operations, most GPU-based linear algebra packages can be applied directly. Our implementation is written in MATLAB using the Jacket GPU library (AccelerEyes). The main tree building uses a standard recursive subdivision algorithm controlled by the host CPU, while the tree splitting algorithm (Algorithm 1) is accelerated on the GPU. Experimental results are presented in Figure 1 and 4. These results were obtained on a PC with NVIDIA 470 GTX graphics card (1.2GB GPU memory) and an Intel Core-i7 3.0GHz CPU with 8 hyperthreads.

Figure 1 lists the computation time using the GPU and CPU implementations. The timing is averaged over 15 runs per test. The GPU implementation generally achieves $5 \sim 12\times$ speedup compared to the CPU version, which is also written in MATLAB and multi-threaded. This performance gain is mainly due to the acceleration of matrix-matrix and matrix-vector multiplications, which can easily exploit the GPU's parallel computation power. Note that the speedup for the KDDCup04 dataset is moderate, mainly because of the small vector size. We generally obtain higher performance gain on higher dimensional datasets, because they can better utilize available GPU resources.

For comparison we have also included a CPU PCA tree implementation using MATLAB's `svds` routine. As seen from the table, the CPU PCA trees are $4 \sim 6$ times slower than the CPU APD trees, which are in turn $5 \sim 12$ times slower than the GPU APD trees. We did not include a GPU PCA tree because the Jacket library only provides a full `svd` routine but not `svds`. As a result, the GPU PCA tree is only moderately faster than the CPU counterpart. This shows that our APD tree algorithm is well-suited for exploiting the GPU, due to its simplicity, at the same time providing comparable quality to PCA trees.

Figure 4 plots the computation times for synthetic data sets containing different numbers of points. Each point is a 512-d vector generated using the same algorithm before. We plot the CPU and GPU timing for each data set and with 0 to 4 power iterations. From this plot we observe that the cost incurred by each additional APD iteration is generally quite small.

## 6. Conclusion

We presented the APD-tree, a new spatial data structure for high-dimensional data. APD-trees use a small number of power iterations to achieve computational efficiency (comparable to RP trees) and high quality (comparable to PCA trees). The approach is insensi-tive to the convergence properties of the power method and is well-suited for GPU computation.

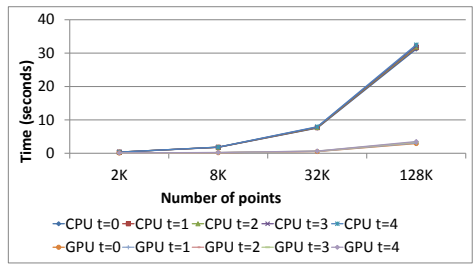

Figure 4. Timing of CPU and GPU APD trees on synthetic datasets with different number of points. Each point is a 512-d vector. Note for both CPU and GPU, the number of power iterations $t$ has little impact on performance.

## Acknowledgments

This work is supported in part by NSF grants CCF-0953754 and CCF-1025120.